\documentclass[preprint,number,12pt]{elsarticle}

\usepackage{amssymb}
\usepackage{amsmath}
\usepackage{graphicx}
\usepackage{float}

\journal{Manufacturing Letters}

\begin{document}

\begin{frontmatter}

\title{A Robotic System for Automated Manufacturing of Dielectric Elastomer Actuators}

\author[uconn]{Van Remenar}
\author[uconn]{Anatol Gogoj}
\author[uconn,uoft]{Ang Li}
\author[uconn]{Tomas Verardi}
\author[uconn]{Matei Mandoiu}
\author[uconn]{Victor Jimenez-Santiago}
\author[uconn,uoft]{Mihai Duduta\corref{cor1}}
\ead{duduta@mie.utoronto.ca} 
\cortext[cor1]{Corresponding author}

\affiliation[uconn]{organization={University of Connecticut},
            addressline={191 Auditorium Road},
            city={Storrs},
            postcode={06269},
            state={CT},
            country={USA}}

\affiliation[uoft]{organization={University of Toronto},
            addressline={5 King's College Road},
            city={Toronto},
            postcode={M5A 3G6},
            state={ON},
            country={Canada}}

\begin{abstract}
This letter presents an automated robotic manufacturing system for soft capacitors which operate as actuators and sensors. Emphasis is placed on the two processes that most directly govern device quality, dielectric layer formation by spin coating and carbon nanotube (CNT) electrode application by stamping. Twenty multilayer DEAs, each comprising 12 dielectric layers with a mean thickness of 55.37 $\pm$ 2.04 $\mu$m and 11 alternating CNT electrodes, were fabricated reducing total process time by 14.2\% and removing the operator from 56.1\% of it



\end{abstract}

\begin{graphicalabstract}
\includegraphics[width=\linewidth]{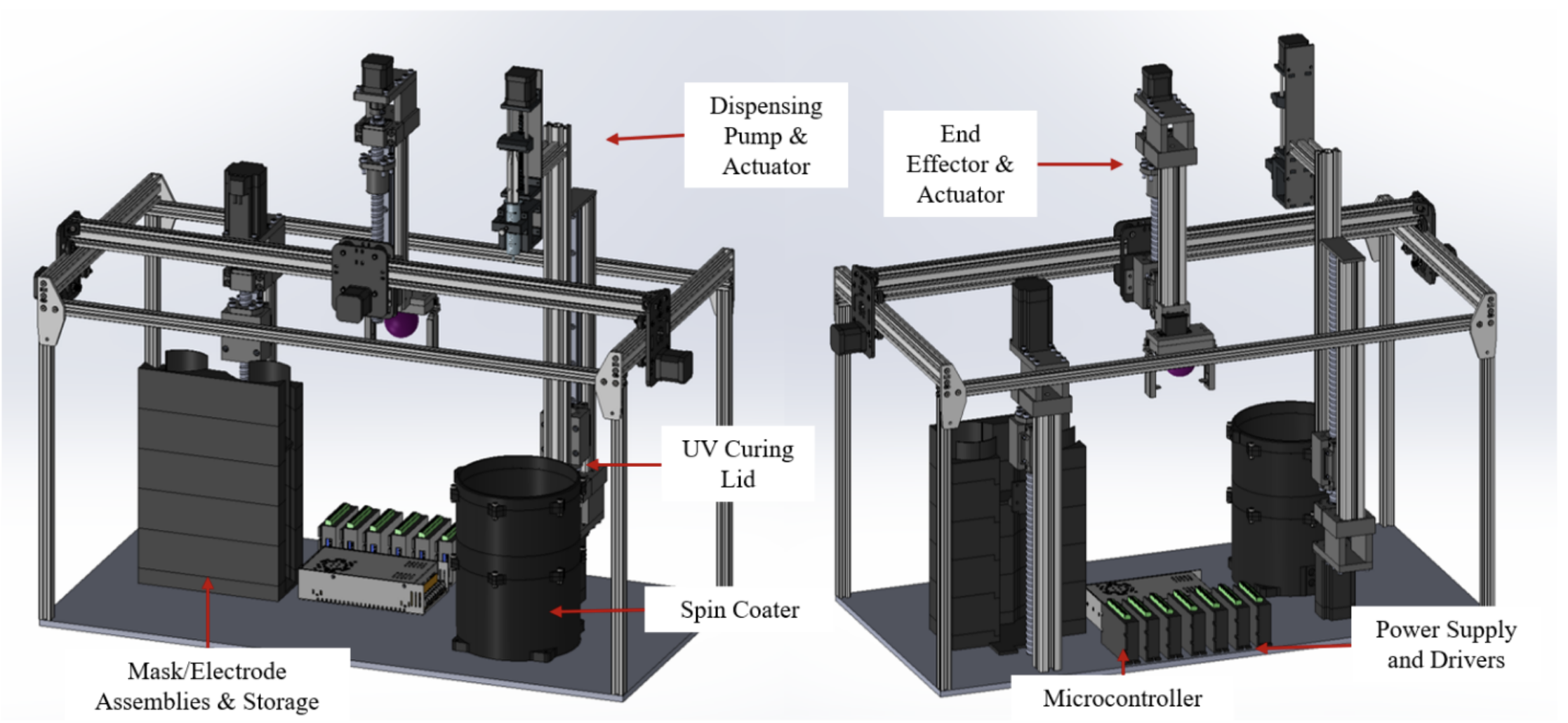}
\end{graphicalabstract}

\begin{highlights}
\item Fully automated platform builds multilayer DEAs on a spin-coater build plate
\item Custom spin coater with rotational homing preserves layer-to-layer alignment
\item Stamping transfers 95.29 $\pm$ 1.47\% of patterned CNT electrode material
\item Combined mask--filter cartridge merges patterning and CNT loading into one step
\item Twenty 12-layer DEAs show consistent displacement and strain across the batch
\end{highlights}


\end{frontmatter}

\clearpage
\section{Introduction}
\label{sec:intro}
Manufacturing multilayer soft-material devices (\textit{e.g.} compliant actuators, stretchable sensors) is limited  by the fabrication process. Dielectric elastomer actuators (DEAs) are a representative case: compliant electrostatic transducers that produce large in-plane strains when a voltage is applied across a soft dielectric membrane coated with stretchable electrodes \cite{pelrine2000,ohalloran2008}. Their low mass and muscle-like mechanics suit soft robotics, wearable devices, and adaptive structures \cite{gu2017,hajiesmaili2021,brochu2010,guo2021}. Because actuation stress scales as $\epsilon E^2$ and the resulting thickness strain inversely with the elastomer modulus \cite{hajiesmaili2021,suo2008,lu2020}, device performance is acutely sensitive to dielectric layer thickness, electrode uniformity, and interfacial quality — properties set almost entirely during fabrication


Despite substantial progress in materials and device concepts, multilayer DEA fabrication remains largely a laboratory craft. Typical process chains combine solution casting or spin coating of dielectric films, stencil- or mask-based electrode deposition, curing, and hand assembly, with an operator handling soft, thin films between every step \cite{ohalloran2008,rosset2016,kruger2023}. This manual dependence introduces device-to-device variability, limits throughput, and makes it difficult to attribute performance differences to design rather than to fabrication. Automated routes such as soft-lithography techniques \cite{corbaci2018}, spray deposition \cite{araromi2011,sprayspin2023}, multilayer stack fabrication \cite{ma2024}, and direct printing of complete actuators \cite{poulin2015,palmic2022} improve individual steps, and compliant carbon nanotube (CNT) electrodes have been applied by brushing \cite{shigemune2018}. However, each of these automates a single operation, leaving manual transfer, alignment, and handling between the steps. 

This letter reports an automated robotic platform that closes this gap by integrating dispensing, spin coating, curing, and electrode application into one programmable manufacturing cycle. The discussion centers on the spin-coating module and the stamping mechanism, the two subsystems that most directly determine device quality, and on batch-level validation and process timing across twenty fabricated devices.

\section{Automated manufacturing platform}
\label{sec:platform}
\begin{figure}[H]
\centering
\includegraphics[width=\linewidth]{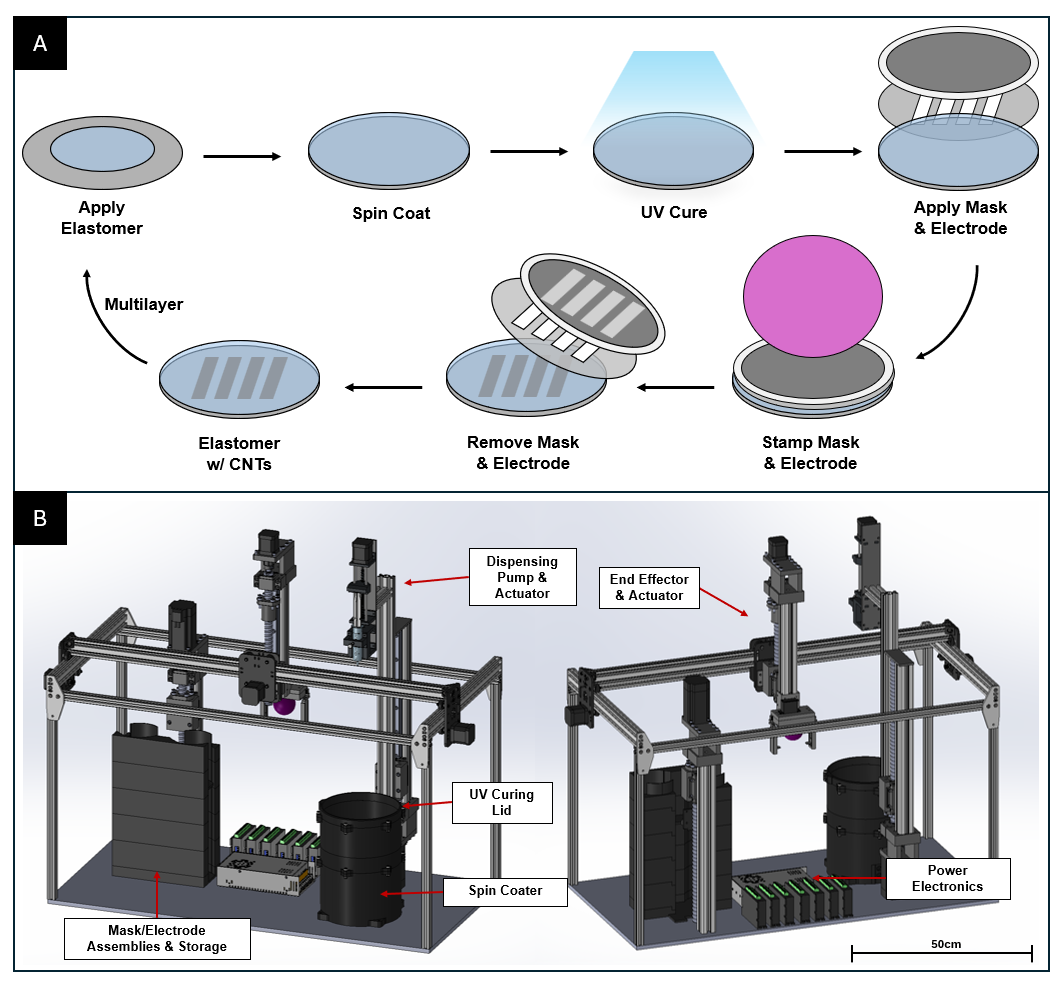}
\caption{Manufacturing overview (A) Steps required to produce a layer of elastomer, stamped with carbon nanotubes at specific locations. (B) CAD model of the manufacturing system including components for dispensing liquid elastomer precursors, spin coating, UV curing, and electrode stamping through a mask. The complete platform operates without manual intervention between steps and occupies a 1 m $\times$ 0.5 m footprint.}
\label{fig:system}
\end{figure}

For each layer, the platform dispenses uncured dielectric, spin coats it, cures it under UV light, retrieves an electrode assembly, and stamps the electrode, repeating this five-step cycle until the device is complete (Fig.~\ref{fig:system}). Every subsystem was designed around this full cycle.

The device is built directly on the spin coater's substrate, which serves as a stationary build plate for the entire cycle. Rather than shuttling the delicate part between stations, a belt-driven Cartesian gantry brings each process to the device, homing against limit switches each cycle so that positional errors do not accumulate. A stepper-driven syringe pump on a retractable lead-screw actuator deposits a controlled volume of uncured elastomer. The custom spin-coating module, described in Section~\ref{sec:spin}, then forms and cures the dielectric film. Electrode application, described in Section~\ref{sec:stamp}, draws pre-loaded assemblies from a 60-assembly storage elevator whose two stacks alternate lead orientation so that successive electrodes connect correctly to power and ground; a single gantry-mounted gripper both retrieves each assembly and performs the stamping stroke.

Marlin-derived firmware on a primary microcontroller centralizes control, with a secondary controller managing the closed-loop spin coater. The platform accommodates a UV-curable elastomers, with relay-controlled argon delivery for formulations that require inert-atmosphere.

\section{Dielectric layer formation by spin coating}
\label{sec:spin}
Spin coating was selected because of the compact footprint compared to blade coating. Automated multilayer fabrication imposes a key requirement: it must return the substrate to a known angular position after every spin so that the CNT electrode pattern stays consistently oriented from layer to layer. Commercial systems lack rotational homing and rely on proprietary control software that prevents integration.

\begin{figure}[H]
\centering
\includegraphics[width=\linewidth]{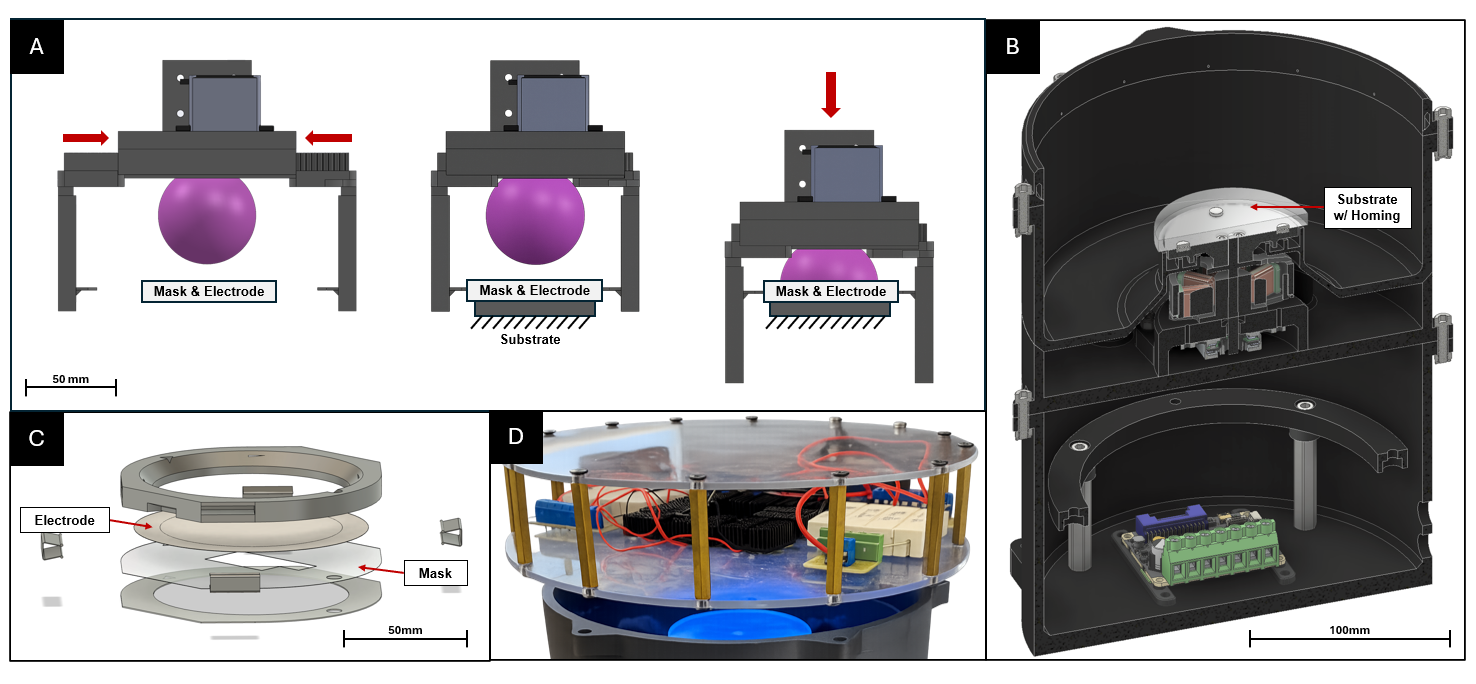}
\caption{Quality-governing subsystems. (A) Gripping and stamping through a soft elastomer pad. (B) Custom spin coater capable of homing to a specific angle orientation. (C) Mask--filter cartridge, pre-aligned and assembled. (D) Spin coater lid, with UV-curing LEDs. Structural components are printed in carbon-fiber-reinforced PETG for stiffness and heat resistance.}
\label{fig:subsystems}
\end{figure}

A custom spin coater was made to include: a rotor driven by a dual-shaft motor paired with an encoder and a closed-loop controller, providing speed regulation during spinning and positional feedback for rotational homing (Fig.~\ref{fig:subsystems}B). The substrate sits on a magnetic mount that supports automated placement and removal. A magnetic catch makes the module removable for service and incorporates dispersion ports for argon in oxygen-sensitive curing. 

Curing is integrated into the coating station through a custom lid carrying UV LEDs positioned to distribute light uniformly across the substrate. A servomotor rotates the lid 90$^\circ$ between a closed position, in which it contains the spinning elastomer and cures the film immediately after coating, and an open position that clears the station. Curing in place eliminates a substrate transfer that would otherwise risk disturbing the uncured layer (Fig.~\ref{fig:subsystems}D).

\section{CNT electrode application by stamping}
\label{sec:stamp}
Stamping is preferred to spraying because stamping allows alternative electrodes to be incorporated without needing to develop spray-compatible inks. Conventional workflows treat electrode patterning and conductive-material loading as separate operations; here they are merged into a single four-layer stamping assembly \cite{duduta2025manufacturing,duduta2016} (Fig.~\ref{fig:subsystems}C), in which the laser-cut mask carrying the electrode pattern and the CNT-loaded filter are sandwiched between a 0.4 mm aluminum backing sheet with a central opening and a rigid 3D-printed top plate. Each cartridge is prepared off-line using alignment holes and corner clamps.

To retrieve an assembly, the gripper closes two servo-driven arms around it (Fig.~\ref{fig:subsystems}A). Passive nail features at the fingertips carry the assembly during transport and slide relative to it during pressing, so it remains seated on the substrate as the gripper continues its downward stroke. A lead-screw actuator drives the stroke, providing the controlled descent speed and consistent force that the transfer step requires, and pressure is delivered through a compliant pad that distributes load across the electrode area in the manner of a pad-printing interface.

Transfer efficiency---the fraction of patterned CNT material moved from filter to dielectric---was 95.29 $\pm$ 1.47\% across ten electrode layers spanning both power and ground orientations (Fig.~\ref{fig:results}B).

\section{Batch fabrication and actuation performance}
\label{sec:results}
Twenty multilayer DEAs were fabricated by the platform under identical conditions. Each device comprises 12 dielectric layers and 11 CNT electrodes, alternating between power and ground. The mean dielectric layer thickness across the batch, measured from scanning electron microscopy (SEM) cross-sections, was 55.37 $\pm$ 2.04 $\mu$m, a coefficient of variation below 4\%, indicating that the dispensing, spin-coating, and curing sequence reproduces film-formation conditions from device to device (Fig.~\ref{fig:sem}A). All reported uncertainties are one standard deviation across the twenty-device batch.

\begin{figure}[H]
\centering
\includegraphics[width=\linewidth]{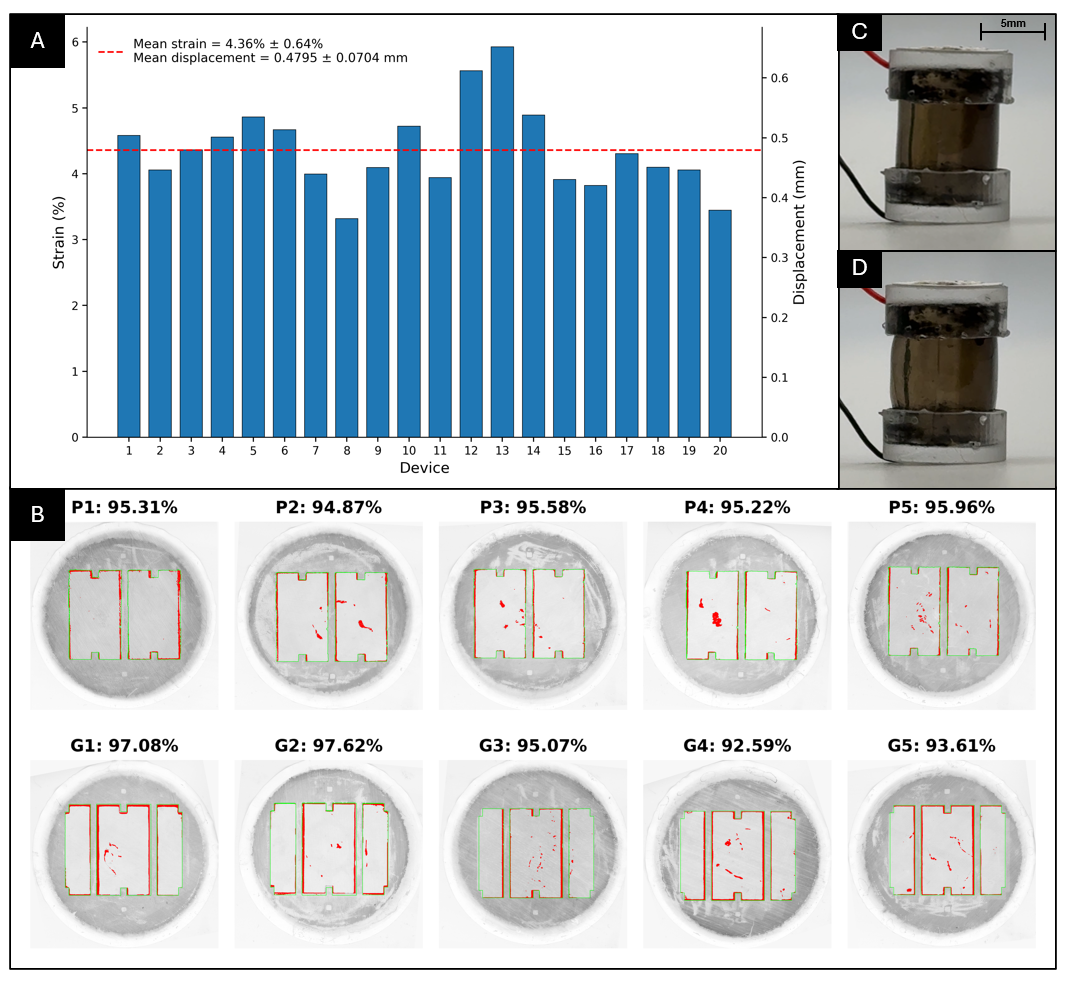}
\caption{Characterization of devices made via the system. (A) Displacement and strain of 20 individual actuators made via the platform. (B) Quantified transfer efficiency of CNTs onto elastomer via stamping. (C) Example of a single rolled DEA, at rest (C) and actuated (D).}
\label{fig:results}
\end{figure}

\begin{figure}[H]
\centering
\includegraphics[width=\linewidth]{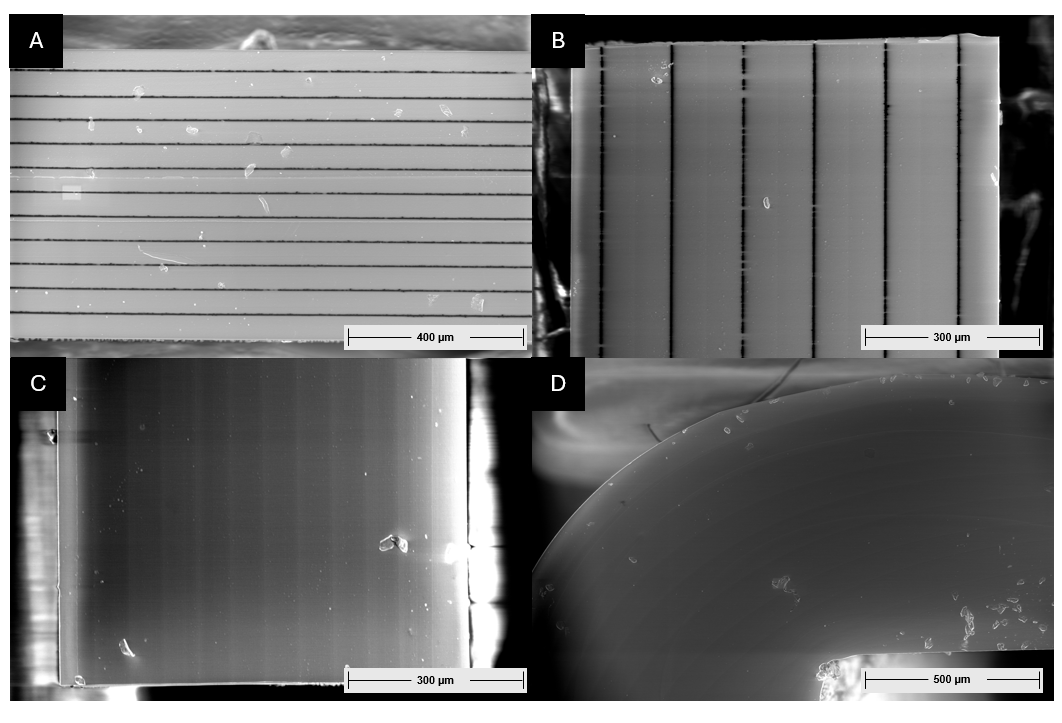}
\caption{Imaging under scanning electron microscopy of different cross-sections in the material. (A) Active region - showing alternating power and ground CNT electrodes. (B) Connection region - showing only power electrodes every second layer of elastomer. (C) Inactive  region - with no CNTs present. (D) Substrate edge - where surface tension caused defects are expected.}
\label{fig:sem}
\end{figure}

After automated fabrication, each planar stack was released from the build plate, cut to a 1.5 cm $\times$ 5 cm footprint, and rolled about an 8-mm mandrel into a cylindrical actuator (Fig. 3C,D); this is the configuration in which all actuation data below were collected. Axial displacement was measured by laser displacement sensor at an applied field of 35 V/$\mu$m (1.9 kV across the 55 $\mu$m mean layer thickness). The batch exhibited a mean displacement of 0.4795 $\pm$ 0.0704 mm, corresponding to a mean actuation strain of 4.36 $\pm$ 0.64\% (Fig.~\ref{fig:results}A). Because every device was produced by the same programmed cycle with no operator handling between steps, the observed spread reflects the intrinsic repeatability of the integrated process.

Process time was measured over a complete build. The fully manual process required 731 s to produce one cut and assembled device, against 627 s for the hybrid automated process: a reduction of 14.2\%, or 104 s per device. Automated operation accounted for 352 s of the hybrid process, so approximately 56.1\% of it proceeded without operator involvement. 


\section{Conclusions}
\label{sec:concl}
An automated robotic system for multilayer DEA manufacturing has been demonstrated, built entirely from subsystems custom designed around the integrated fabrication cycle. A custom spin coater and a gripper-driven stamping mechanism form each dielectric layer and apply each electrode in place; twenty 12-layer devices showed consistent layer thickness and actuation performance while cutting total process time by 14.2\%. Ongoing work addresses pressure uniformity in fine electrode features, broader material validation, and in-process sensing for closed-loop quality control \cite{perez2016,peng2023,inayathullah2025}.

\section*{Funding}
This research was supported by grants from the U.S. Army Ground Vehicles Systems Center (GVSC) through the Digital Design Research, Analysis, and Manufacturing (DREAM) Center at the University of Connecticut (VR, AG, VJS, MD). 

\section*{CRediT authorship contribution statement}
\textbf{Van Remenar:} Conceptualization, Methodology, Investigation, Data curation, Writing -- original draft. \textbf{Anatol Gogoj:} Investigation. \textbf{Ang Li:} Investigation. \textbf{Tomas Verardi:} Investigation. \textbf{Matei Mandoiu:} Investigation. \textbf{Victor Jimenez-Santiago:} Investigation. \textbf{Mihai Duduta:} Conceptualization, Supervision, Resources, Writing -- review \& editing.

\section*{Declaration of competing interest}
The authors declare that they have no known competing financial interests or personal relationships that could have appeared to influence the work reported in this paper.

\section*{Data availability}
Data is available from the authors upon reasonable request.

\end{document}